\documentclass[9pt, conference, compsocconf]{IEEEtran}
\usepackage{cite}

\ifCLASSINFOpdf
   \usepackage[pdftex]{graphicx}
\else
\fi
\usepackage[cmex10]{amsmath}
\usepackage[tight,footnotesize]{subfigure}
\usepackage{url}

\usepackage{booktabs}
\usepackage{multirow}
\usepackage[bookmarks=false]{hyperref}

\begin{document}
%
\title{Addressing the Cold-Start Problem in Outfit Recommendation Using\\ 
Visual Preference Modelling}


\author{\IEEEauthorblockN{Dhruv Verma, Kshitij Gulati and Rajiv Ratn Shah}
\IEEEauthorblockA{Indraprastha Institute of Information Technology, Delhi, India\\
\{dhruv17046, kshitij17162 and rajivratn\}@iiitd.ac.in
}}


%


\maketitle

\begin{abstract}
With the global transformation of the fashion industry and a rise in the demand for fashion items worldwide, the need for an effectual fashion recommendation has never been more. Despite various cutting-edge solutions proposed in the past for personalising fashion recommendation, the technology is still limited by its poor performance on new entities, i.e. the cold-start problem. In this paper, we attempt to address the cold-start problem for new users, by leveraging a novel \textit{visual preference modelling} approach on a small set of input images. We demonstrate the use of our approach with \textit{feature-weighted clustering} to personalise occasion-oriented outfit recommendation. Quantitatively, our results show that the proposed visual preference modelling approach outperforms state of the art in terms of clothing attribute prediction. Qualitatively, through a pilot study, we demonstrate the efficacy of our system to provide diverse and personalised recommendations in cold-start scenarios.
\end{abstract}


%
\IEEEpeerreviewmaketitle

\section{Introduction}
\noindent
With the growing influence of online social media, people have a strong desire to remain connected sharing almost every aspect of their daily life, from the delicious food they are having, TV shows they are binge-watching, places they are visiting, to what outfits they are wearing. Since an individual’s dressing style speaks volumes about his/her personality, it has given increased importance to the appropriateness of an outfit in a particular context. We always try to portray a better version of ourselves and in this case fashion knowledge not only helps us look better, but it speaks to the physiological needs and caters to the demands of social events and activities. With this increased consciousness, e-commerce for fashion is booming, but consumers still face issues while selecting fashion outfits that are appropriate in context and suit their liking. 

\begin{figure}[t]
    \centering
    \includegraphics[width=0.78\columnwidth]{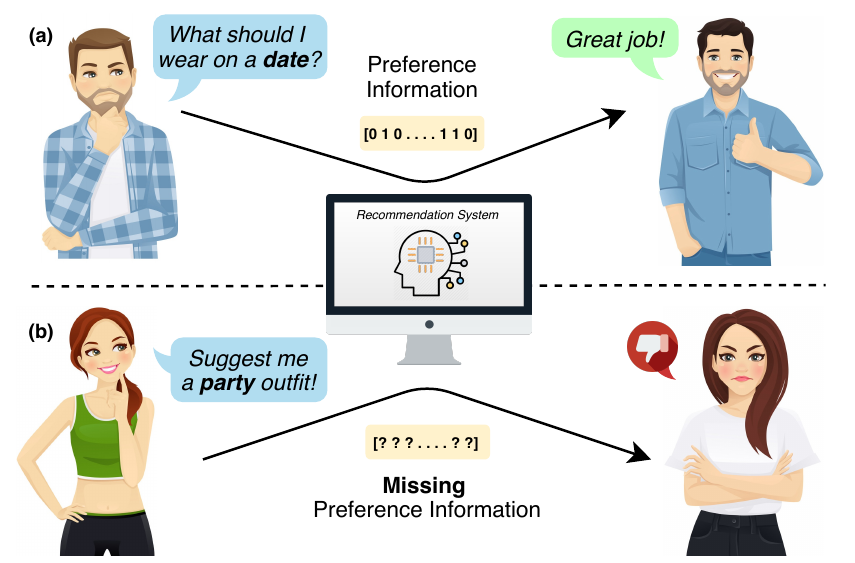}
    \caption{An illustration of the cold-start problem in outfit recommendation for new users. (a) represents a profiled user, whose preferences are known to the system, while (b) represents a new user.}
    \label{fig:cold-start}
\end{figure}

In the past, various studies have attempted to tackle different fashion-related issues like fashion retrieval \cite{liao_interpretable_2018, mustaffa_dress_2019}, clothing recognition and classification \cite{kalantidis_getting_2013,hidayati_clothing_2012}, outfit recommendation \cite{lin_outfitnet_2020,chen_personalized_2019}, etc. When we talk about outfit recommendation, it cannot be generic, since user preference is inherently subjective. User preference is a reﬂection of age, occupation, culture, place of living, etc. Therefore, personalisation is essential because it ensures outfit recommendation is in line with the users’ personal fashion taste and incorporates users’ likes and dislikes from a variety of perspectives. Personalised fashion recommendation has gained a lot of attention in recent work \cite{hu_collaborative_2015, chen_pog_2019,yin_enhancing_2019, chen_personalized_2019}. However, most of them fail to address the cold-start problem \cite{lika_facing_2014} for new users (See Figure \ref{fig:cold-start}) and solving this issue forms the basis for our approach. 

Fashion knowledge can best be represented in terms of person, occasion and clothing \cite{ma_who_2019}. Presently, various fashion datasets comprising DeepFashion \cite{liu_deepfashion_2016}, Polyvore~\cite{han_learning_2017}, FashionVC~\cite{song_neurostylist_2017}, etc. have been collected. However, none of these has adequate representation for user-context as well as occasion. Consequently, we curate a high-quality dataset, constituting images scraped from Instagram and Pinterest, representing a diverse variety of occasion-oriented fashion knowledge.
The optimal recommendation for a particular occasion scenario would essentially be suitable for that occasion while keeping the fashion elements or styles as per the user’s liking. For example, a man attending a conference would be recommended to dress up in formal attire, say a suit. However, elements of the outfit like colour, texture, style, etc. can be modulated as per his preference. In this paper, we attempt to address the cold-start problem for new users by modelling their preference visually. By taking a few images of the users depicting their fashion taste as an input, we aim to extract rich fashion concepts and model these extracted concepts as preferences to the recommender for cold-start scenarios.
\par
We experiment with various pre-existing techniques to extract fashion concepts from the users’ images. Our model predicts clothing categories with an accuracy of \textit{73.5\%}, F1 score of \textit{0.72} and attributes with an accuracy of \textit{89.1\%}, F1 score of \textit{0.88}. Our model outperforms state of the art in attribute classification and has comparable performance for category classification. We further evaluate our system by conducting a pilot study on ten subjects and find that most users find recommendations with visual preference modelling more relevant as opposed to recommendations without~it.

The main contributions of this work are as follows:
\begin{itemize}
    \item We contribute a fashion knowledge dataset and demonstrate the efficacy of multi-task learning in terms of capturing fashion concepts from a wide variety of images.
    
    \item We propose a novel visual preference modelling approach with feature-weighted clustering for personalising outfit recommendation in cold-start scenarios\footnote{Source code and dataset: \textit{\url{https://github.com/Dhruv-Verma/fashonist}}.}.
\end{itemize}

\section{\textbf{Related Work}}
\noindent
The main objective of this paper is to address the cold-start problem in outfit recommendation. Hence, we primarily discuss recent developments in fashion concept prediction and personalised outfit recommendation in this section.       

\subsection{Fashion Concept Prediction}
Fashion outfits contain various high-level concepts like style, season, occasion and environment, which can be inferred from design elements like shape, colour, cloth, texture, etc. \cite{sorger_fundamentals_2012}. Studies conducted in the past \cite{vaccaro_elements_2016} suggest that low-level elements of fashion like clothing categories and fine-grained attributes can be translated to these high-level fashion concepts. To this end, various computer-vision based approaches have been explored for representing clothing elements from images. With the availability of rich fashion databases like DeepFashion \cite{liu_deepfashion_2016}, ModaNet \cite{zheng_modanet_2019}, etc., deep neural network-based approaches have risen to new heights in terms of performance. Liu et al. \cite{liu_deepfashion_2016} leveraged a branched CNN architecture, \textit{FashionNet}, which learns clothing features by jointly predicting clothing attributes and landmarks. Ma et al.~\cite{ma_who_2019} proposed a contextualized fashion concept learning module using Bi-LSTMs integrated with a CNN backbone. Yan et al. \cite{yan_multi-task_2019} proposed  \textit{StyleNet} which used multi-task representation learning for creating clothing representations accommodating different fashion concepts like seasons, garments and styles. While some of these approaches were successful in capturing the dependencies and correlation between different fashion elements and concepts, their results could not show significant improvement in predicting fine-grained clothing attributes. Since multi-task learning has shown promising results in many applications \cite{zhang_survey_2018}, we are motivated to leverage the information-sharing capabilities of multi-task learning in the domain of fashion concept prediction while modelling the dependencies between clothing categories and attributes. With this approach, we observe a significant improvement in performance in comparison with existing work while keeping the computational complexity low. This is attributed to our elemental approach towards clothing representation along with dependence modelling.

\subsection{Cold-Start Problem in Outfit Recommendation}
In recent work, various successful attempts towards personalising outfit recommendation have been made, viz; tensor factorization \cite{hu_collaborative_2015}, visual compatibility modelling \cite{chen_pog_2019,yin_enhancing_2019}, and multimodal attention-based networks \cite{chen_personalized_2019,lin_outfitnet_2020}. However, all of these works are limited in scope by their inability to address the cold-start problem for outfit recommendation. Only a small body of work has focussed on tackling this problem. Bracher et al. \cite{bracher_fashion_2016} proposed novel latent space representation for fashion items called \textit{FashionDNA}. \cite{piazza_emotions_2017,sun_learning_2013} used visual representations incorporating categories and styles for clothing items. These methods attempted to address cold-start in terms of new items for recommendation but overlooked the new user aspect. Piazza et al. \cite{piazza_emotions_2017} evaluated the predictive power of users’ affective information for mitigating the cold-start problem for new users. Their findings suggest that mood information positively influences the quality of predictions. However, this improvement was not significant. We, therefore, find an opportunity to address this problem via a practical approach. We exploit fashion concept prediction for inferring preferences from a small set of input images for a new user.

\section{\textbf{Dataset}}
\noindent
Social media platforms offer rich user-centric data with images of users across the world, depicting diverse real-world fashion knowledge with natural backgrounds. These images usually contain indications to various occasions such as conferences, weddings and parties. We contribute a dataset which comprises of 2893 high-quality images scraped from Instagram and Pinterest. We collect images for seven types of occasions, namely \emph{Travel, Sports, Dating, Wedding, Party, Office} and \emph{Prom}, by using the presently trending hashtags and keywords associated with these occasions. For example, we used \emph{\#sportsoutfit} for Sports and \emph{\#officewear} for Office. We ensure that images extracted are balanced in terms of gender and are easily distinguishable and diverse in terms of various fashion concepts. 

\begin{figure}[b]
    \centering
    \includegraphics[width=1\columnwidth]{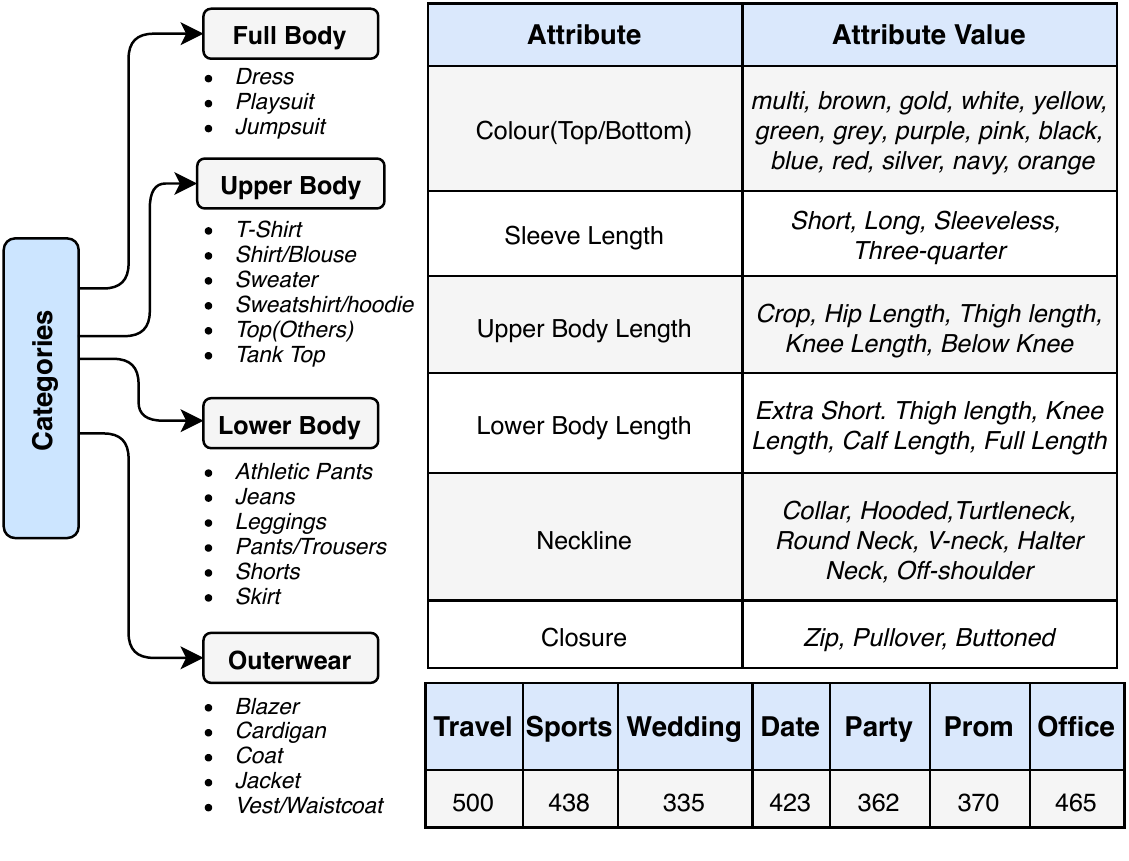}
    \caption{Our Dataset, with 20 categories, 39 attributes across 7 occasions}
    \label{fig:annotations}
\end{figure}

After extracting these images, we perform manual filtering to rule out all possible noisy images which do not adequately represent fashion knowledge and have poor quality, low brightness, side angle shots and selfies. We use a pre-trained CNN-based model \cite{levi_age_2015} for age and gender classification of each image.
We define a set of categories and attributes to best represent the fashion knowledge in terms of the clothing aspects for each image, as shown in Figure~\ref{fig:annotations}. Each category defines a particular region of clothing (e.g. t-shirt: upper body, athletic pants: lower body) and each attribute further characterises a particular category (e.g. sleeve length, neckline: t-shirt, lower body length: athletic pants).  

Two fashion experts annotate each image following our annotation scheme. For evaluating the reliability of annotations, an inter-annotator agreement is calculated using kappa statistics. In case of disagreement for a particular annotation, a third fashion expert breaks the tie. As a general guideline, a kappa value above 0.6 is considered adequate and above 0.8 is considered almost perfect \cite{nowakuger}. We find the inter-annotator agreement for both tasks to be reasonably high with a kappa value of \emph{0.81} for category annotation and a kappa value of \emph{0.68} for attribute annotation. As a result, we achieve our fashion knowledge database which describes each image in terms of the \textit{person} (Gender and Age), \textit{clothing} (Categories and Attributes) and \textit{occasion}. 

\begin{figure*}
    \centering
    \includegraphics[width=0.95\linewidth]{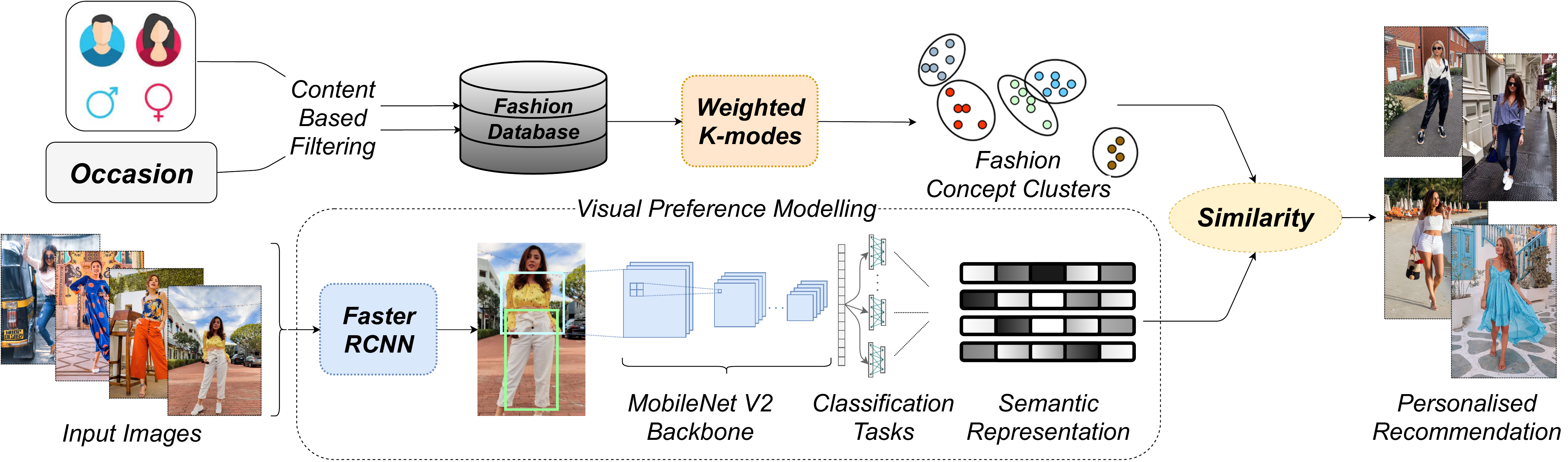}
    \caption{The proposed pipeline for personalised outfit recommendation in cold-start scenarios}
    \label{fig:pipeline}
\end{figure*}

\section{\textbf{Methodology}}
\noindent
Our objective is to demonstrate the use of fashion concept prediction to model the preferences of new users given a small set of images as input. We expect the users to provide images that depict a diverse range of fashion choices highlighting their likes and dislikes towards various fashion elements. The first portion of our work is focused on the preference modelling of visual fashion elements like clothing categories and attributes. In the second portion, we discuss our strategy for incorporating the visual preferences into outfit recommendation. (See Figure \ref{fig:pipeline})

\subsection{Visual Preference Modelling}
\label{subsec:vpm}
Fashion outfits are a combination of different clothing items (e.g. tops and bottoms or full body dresses). To extract fashion concepts with efficacy, we first need to represent different fashion items in terms of regions of interest. We use Faster RCNN (with a ResNet-50 backbone) \cite{faster-rcnn}, which is a state of the art technique for real-time object detection. We fine-tuned the Faster RCNN (pre-trained on the MS-COCO dataset \cite{lin_microsoft_2015}) on the ModaNet dataset \cite{zheng_modanet_2019} to ensure that it only detects relevant fashion elements as regions of interest. 
\par
These regions of interest are unlabelled and cannot directly be used to represent visual fashion knowledge. Therefore, we need to transform them into a semantic space representation which can encode the visual fashion knowledge in terms of different categories and attributes. Since our objective is to recommend outfits from our curated dataset, the semantic space representation of the users’ images should also incline with the labelled fashion knowledge in our dataset. Hence, for a given image $I$, we have a set of \textit{clothing regions} $\{r_1, r_2,..r_i..,r_n\}$, the goal is to predict \textit{category label} $c_{r_i} \in \{c_1,c_2,...,c_m\}$ and a set of \textit{attribute values} $\{a^1_{r_i},a^2_{r_i},..a^j_{r_i},.,a^p_{r_i}\}$ where $a^j_{r_i} \in$ \textit{attribute types} $\{a^j_{1},a^j_{2},...,a^j_{x}\}$ for each $r_i$. Hence, each region of interest is characterized by a clothing category and attribute values semantically similar to our labeled dataset. Thereafter, the representations of individual regions of interest are pooled together to form vector $v_I$ for every image $I$ in the given collection of users' images.
\par
Since clothing categories and attribute values are not independent entities, there exist dependencies between them. In order to learn the hidden representations capable of accommodating individual features as well as dependencies, we make use of multi-task learning. We use a multi-task classifier with a MobileNet V2 \cite{howard_mobilenets_2017} backbone. Different classification heads are responsible for optimizing the category and attribute prediction tasks. The minimization objective function can be described as in Equation~\ref{eqn:loss}.\textbf{
\begin{equation}
    \label{eqn:loss}
    L_{cumulative} = \lambda_1 L_{category}  + \lambda_2 \sum L_{attribute} 
\end{equation}}
$\lambda1$ and $\lambda2$ are the weights for category and attribute losses respectively. The cumulative loss function ensures that the shared hidden representation incorporates both individual as well as dependency-based features. The MobileNet backbone is initially trained on the large DeepFashion \cite{liu_deepfashion_2016} dataset to learn various high-level feature maps relevant for fashion domain tasks. Finally, this backbone with multiple classification heads is fine-tuned on our dataset.

\subsection{Personalised Outfit Recommendation}
After extracting the vector representation (as described in Section \ref{subsec:vpm}) for the given input images, our goal is to fuse this extracted preference knowledge into occasion-oriented outfit recommendation. Within the boundaries of a particular occasion, relevant outfits may represent various high-level fashion concepts like style, patterns, seasons, etc. For example, male outfits belonging to the occasion \textit{Office} may form subgroups based on styles present, viz; a style subgroup may have a more formal appeal (e.g. suits) while another may depict a more casual style like a sweater with chinos. Similarly, the presence of outerwear and attribute values like sleeve length and neckline could lead to the formation of subgroups depicting different seasons.    
\par
After applying content-based filters for gender and occasion on our dataset, we retrieve the relevant outfit images. We need to extract the high-level concept subgroups from this set. Since the outfits in this set are labelled with appropriate category and attribute information, we use an unsupervised learning approach, a modified version of the \textit{k-modes} clustering algorithm \cite{chaturvedi2001k} for forming the concept clusters. The modifications include a weighted dissimilarity measure instead of the simple hamming distance (as described in \cite{saranganayagi_improved_2009}) and different weights $\alpha$ and $\beta$ for the category and attribute related features respectively (Refer Equation 2). For identifying the optimal number of clusters to be formed for each subgroup, we use the metric \textit{silhouette index} \cite{aranganayagi_clustering_2007}.
\begin{equation}
    D_m (X,Y) = \alpha\sum_{j \in cat}\delta (x_j,y_j) + \beta\sum_{j \in attr}\delta (x_j,y_j)
\end{equation}
After forming the concept clusters, a trivial approach is to draw an equal number of samples from each cluster for the final recommendation. While this may ensure greater diversity in the recommendation, it does not account for incorporating the users' preferences in any way. 
Therefore, we weight each cluster by calculating the similarity between the semantic representations of outfits in the input images and the concept clusters (semantic representation of centroids used). The number of samples drawn for the recommendation is decided based on the weights assigned to each cluster. For example, if some images given by a female user represent her liking for sleeveless dresses (category: dress, sleeve length: sleeveless), the recommended results are likely to have a greater number of outfits reflecting these features. By doing so, we ensure the diversification of results in addition to the incorporation of users' preferences in a cold-start scenario. 

\begin{table}[h]
\centering
\caption{Performance comparison of different CNN architectures}
\label{table: compare_mtl}
\begin{tabular}{@{}lccccc@{}}
\toprule
\multicolumn{1}{c}{\multirow{2}{*}{\textbf{\begin{tabular}[c]{@{}c@{}}Backbone \\ (Configuration)\end{tabular}}}} & \multirow{2}{*}{\textbf{Task}} & \multicolumn{2}{c}{\textbf{Categories}} & \multicolumn{2}{c}{\textbf{Attributes}} \\
\multicolumn{1}{c}{}                                                                                              &                                & \textbf{Acc.}       & \textbf{F1}       & \textbf{Acc.}      & \textbf{F1}        \\ \midrule
\multirow{3}{*}{\begin{tabular}[c]{@{}l@{}}ResNet-18 \\ (fine-tuned 1 block)\end{tabular}}                        & category                  & 0.68                & 0.65              & -                  & -                  \\
                                                                                                                  & attribute                 & -                   & -                 & 0.73               & 0.71               \\
                                                                                                                  & multi-task                     & 0.69                & 0.68              & 0.87               & 0.86               \\ \midrule
\multirow{3}{*}{\begin{tabular}[c]{@{}l@{}}ResNet-50 \\ (fine-tuned 1 block)\end{tabular}}                        & category                  & 0.65                & 0.64              & -                  & -                  \\
                                                                                                                  & attribute                 & -                   & -                 & 0.71               & 0.70               \\
                                                                                                                  & multi-task                     & 0.68                & 0.665             & 0.86               & 0.85               \\ \midrule
\multirow{3}{*}{\textbf{\begin{tabular}[c]{@{}l@{}}MobileNet-V2 \\ (fine-tuned 3 blocks)\end{tabular}}}            & category                  & 0.71                & 0.69              & -                  & -                  \\
                                                                                                                  & attribute                 & -                   & -                 & 0.76               & 0.74               \\
                                                                                                                  & \textbf{multi-task}            & \textbf{0.735}      & \textbf{0.72}     & \textbf{0.89}      & \textbf{0.88}      \\ \bottomrule
\end{tabular}
\end{table}

\vspace{-2mm}
\section{\textbf{Classification Results}}
\noindent
For performing robust fashion concept prediction using multi-task learning, we experimented with different networks as backbones and fine-tuned them with different configurations. In this section, we briefly describe the performance of these network architectures and compare them with baselines (single classification task CNNs) and state of the art. We experimented with various CNN-based backbones like MobileNet V2, ResNet-18 and ResNet-50 for classification of clothing categories and attributes in both single and multi-task configurations. We initially train these networks on DeepFashion \cite{liu_deepfashion_2016} and then on our dataset. We use the \textit{cross-entropy} loss for each classification head, \textit{adam} optimizer with a learning rate of 0.001, iterated for 100 epochs. We find that \textbf{multi-task classifier with MobileNet backbone} (with 3 blocks of fine-tuning, $\lambda_1/\lambda_2 = 2$, $\alpha/\beta = 2$, calculated empirically) outperforms its alternatives, with an accuracy of \textbf{73.5\%} and \textbf{89.1\%} for category and average attribute classification respectively (Refer Table \ref{table: compare_mtl}). 

We also compare the performance of our model with state of the art architectures like FashionKE \cite{ma_who_2019} and FashionNet \cite{liu_deepfashion_2016}. Our model compares well these systems in terms of category classification and outperforms them for attribute classification (Refer Table~\ref{table: compare_sota}). Our results demonstrate the value added by the information-sharing capabilities of multi-task learning.

\begin{table}[t]
\centering
\caption{Performance comparison with state of the art architectures}
\label{table: compare_sota}
\begin{tabular}{@{}lclcl@{}}
\toprule
\multicolumn{1}{c}{\multirow{2}{*}{\textbf{Model}}} & \multicolumn{4}{c}{\textbf{Accuracy}}                                           \\
\multicolumn{1}{c}{}                                & \multicolumn{2}{c}{\textbf{Category}} & \multicolumn{2}{c}{\textbf{Attributes}} \\ \midrule
FashionNet {[}16{]}                                 & \multicolumn{2}{c}{67.33\%}           & \multicolumn{2}{c}{65.6\%}              \\
FashionKE {[}11{]}                                  & \multicolumn{2}{c}{\textbf{73.95\%}}           & \multicolumn{2}{c}{69.59\%}             \\
\textbf{MobileNet MTL (Our Method)}                 & \multicolumn{2}{c}{73.5\%}   & \multicolumn{2}{c}{\textbf{89.1\%}}     \\ \bottomrule
\end{tabular}
\end{table}

\vspace{1mm}
\section{\textbf{Pilot Study and Evaluation}}
\noindent
In order to evaluate our proposed method for personalised outfit recommendation, we conducted a brief pilot study with ten subjects (five male and five female, belonging to an age group of 18 - 35). We asked users to rate each recommendation on a scale of 10 in terms of \textit{relevance} and \textit{diversity}. We ask users to give a total of 10 images displaying their fashion preferences along with the occasion for which they wish to dress up. The system then recommends\footnote{The recommendations follow no particular order or ranking.} the user a set of 10 outfits for that particular occasion both with and without integrating their personal preferences. The users are not aware of whether the recommendations are inclusive of their preferences or not, to eliminate any bias while evaluating. The results of the pilot study (Refer to Table \ref{table: results_pilot_study}) show that users rate the recommendations highly in terms of diversity, affirming the success of the weighted clustering approach. We find that on an average, users find recommendations with visual preference modelling more relevant as opposed to recommendations without it. An important observation is a slight reduction in the diversity score with visual preference modelling. This can be attributed to the increased importance given to fashion concept clusters which are more similar to the users' outfit images, leading to more number of outfits from the same cluster. Overall, the pilot study results confirm that we have been able to optimise outfit recommendation as well as address the cold-start problem with this approach.

\begin{table}[h]
\centering
\caption{Results of the pilot study (Normalised)}
\label{table: results_pilot_study}
\begin{tabular}{@{}lcccc@{}}
\toprule
\multirow{2}{*}{\textbf{Subject}} & \multicolumn{2}{c}{\textbf{Without Preference}}    & \multicolumn{2}{c}{\textbf{With Preference}} \\
                                  & \textbf{Diversity} & \textbf{Relevance} & \textbf{Diversity}          & \textbf{Relevance}         \\ \midrule
Male                              & 0.76               & 0.58               & 0.66                        & 0.73                       \\
Female                            & 0.85               & 0.62               & 0.74                        & 0.81                       \\
\textbf{Average}                  & \textbf{0.81}      & \textbf{0.60}      & \textbf{0.70}               & \textbf{0.77}              \\ \bottomrule
\end{tabular}
\end{table}

\section{\textbf{Conclusion and Future Work}}
\noindent
In this paper, we attempt to address the cold-start problem for new users in occasion-oriented outfit recommendation. We demonstrate the use of the proposed \textit{Visual Preference Modelling} with the  \textit{Feature-Weighted Clustering} method for generating personalised recommendations using a small set of images as input. The results of our preliminary study affirm that our approach is beneficial for improving outfit recommendation in cold-start scenarios. To further assess the performance of our system, we plan to conduct a more extensive evaluation with a larger participant group. We also note that personalisation is significantly dependent on the quality of images provided by the user. The system can hereon be improved by instilling certain high-level fashion concepts like body shape, texture, weather, seasonal variation and cultural dependence of certain clothing styles which can help improve personalisation. We plan on investigating these avenues in the future. 


\clearpage

\bibliographystyle{ieeetr}
\bibliography{references}

\end{document}